# Deep Learning Approach to Predict Hemorrhage in Moyamoya Disease

Meng Zhao[1]*, Yonggang Ma[2]*, Qian Zhang[1], Jizong Zhao[1]

[1]Department of Neurosurgery, Beijing Tiantan Hospital, Capital Medical University, China

[2]Department of NeuroInterventional Surgery, Binzhou Medical University Hospital, China

## Author Note

Correspondence to Jizong Zhao M.D., Department of Neurosurgery, Beijing Tiantan Hospital, Capital Medical University. Email: zhaojz205@163.com

*These authors contributed equally to this work.

# Abstract

**Objective**

Reliable tools to predict moyamoya disease (MMD) patients at risk for hemorrhage could have significant value. The aim of this paper is to develop three machine learning classification algorithms to predict hemorrhage in moyamoya disease.

**Methods**

Clinical data of consecutive MMD patients who were admitted to our hospital between 2009 and 2015 were reviewed. Demographics, clinical, radiographic data were analyzed to develop artificial neural network (ANN), support vector machine (SVM), and random forest models.

**Results**

We extracted 33 parameters, including 11 demographic and 22 radiographic features as input for model development. Of all compared classification results, ANN achieved the highest overall accuracy of 75.7% (95% CI, 68.6%–82.8%), followed by SVM with 69.2% (95% CI, 56.9% – 81.5%) and random forest with 70.0% (95% CI, 57.0% – 83.0%).

**Conclusions**

The proposed ANN framework can be a potential effective tool to predict the possibility of hemorrhage among adult MMD patients based on clinical information and radiographic features.

**Introduction**

Moyamoya disease (MMD) is a chronic cerebrovascular disease characterized by progressive occlusion of the steno-occlusive changes at the terminal intracranial internal carotid arteries[1]. Intracranial hemorrhage and cerebral ischemia are the two main clinical manifestations of MMD. In general, MMD patients presenting with intracranial hemorrhage, which is found in range from 20% to 60% of adult patients, are associated with poor long-term outcomes[2]. The etiology and mechanism of hemorrhage in MMD remain unclear[1,3,4]. Antithrombotic therapy and surgical revascularization, are the standard of care for stroke prevention in patients at high risk of cerebrovascular complications[5]. However, use of antiplatelet agents is controversial[6], because of their poor efficacy in improving blood supply and potentially increased hemorrhagic effect[7]. Thus, early prediction of hemorrhage in moyamoya disease is valuable in clinical practice. Previous studies have described several possible risk factors for cerebral hemorrhage in MMD, including age, gender, hypertension, and combined with cerebral aneurysm. However, there is no reliable clinical tool for prediction of bleeding among moyamoya patients.

Much progress has been made in artificial intelligence (AI) in recent years. The potential of machine learning being applied in medicine has been studied and was reported to have achieved physician-level performance in various tasks[8–10]. Artificial neural network (ANN) is based on a collection of connected units or nodes called artificial neurons, which loosely model the neurons in a biological brain[11]. Support vector machines (SVM) work relatively well when there is a clear margin of separation between classes and are effective in high dimensional spaces[12]. Random forest is a classification algorithm consisting of many decision trees and its performance is boosted via a voting scheme[13]. The base learner of random forest is a decision tree, and random attribute selection is introduced in the training process of the decision tree. It has proven to handle high dimensional data well and is relatively resistant to overfitting

Previous studies reported that machine learning had shown promising results in the diagnosis or outcome prediction in neurological diseases including strokes[14,15]. It could have the potential for hemorrhage prediction in MMD. We describe the development and validation of an automated machine learning system to predict the possibility of hemorrhage among moyamoya patients. Several machine learning models, ANN, random forest, and SVM, were explored. To our knowledge, this is the first machine learning study to predict hemorrhage in MMD.

## Method

**Patients**

Clinical data of consecutive MMD patients who were admitted to our hospital between 2009 and 2015 were reviewed. Details of data collection have been published previously[16]. For algorithm development, the inclusion criteria were as follows: 1) age 18 years or older, 2) the diagnosis of moyamoya disease was made by neurologists or neurosurgeons. We excluded patients who were 1) diagnosed as the quasi-moyamoya disease (moyamoya syndrome), 2) underwent external ventricular drain or revascularization before admitted to our hospital. We defined hemorrhagic MMD (hMMD) as MMD presented with intracerebral hemorrhage (ICH), intraventricular hemorrhage, or Subarachnoid hemorrhage. Patients without hemorrhage occurrence and who were presented with TIA or ischemia were classified as iMMD patients. The study was approved by the Beijing Tiantan Hospital Research Ethics Committee, and informed consent was obtained from all patients.

**Data and parameters extraction**

Clinical information including age at hemorrhage, sex, past medical history, and imaging data utilized for MMD characterization (CT Angiography, MRI, DSA) were obtained. The angiographic stages of MMD were estimated according to the Suzuki angiographic stage classification[17]. Different stages of stenosis severity level in major vessels are classified as normal (0-25%), minor occlusion (25-50%), moderate occlusion (50-75%) and severe occlusion (75-100%). The anterior choroidal artery (AChA) and posterior communicating artery (PCoA) in each hemisphere was recorded as normal or dilated (either is dilated with distal branching). Collateral circulation were categorized into leptomeningeal, duro-pial, and periventricular collaterals according to DSA. The radiographic features were extracted by two neuroradiologists independently. Any disagreement was resolved by a third author.

**ANN development**

Data samples were randomly divided into training (80%) and testing datasets (20%). In order to optimize the algorithms. Hyperparameters, including number of layers, hidden units, learning rate, and batch size, were tuned by grid search. We used a rectified linear unit (ReLU)[18] in the hidden layer as activation function. The degree of error was calculated by the least-squares method, which is named "loss function". Adaptive moment estimation (Adam)[19] was used as the optimization algorithm to minimize the loss function. Necessary gradients were calculated by backward propagation[11]. We examined models with the stratified sampling K-fold cross-validation method (K = 5), which partitions the original training sample into 5 disjoint subsets, uses 4 of those subsets in the training process, and 1 was used for testing validation. We averaged model performance metrics for statistical analysis. The training and testing procedures were implemented by keras and tensorflow with python 3.7. Custom codes for the deployment of the system are available for research purposes from the corresponding author upon request.

**SVM and Random Forest framework development**

SVM is effective in handling both linear and non-linear data. We used the radial basis function (RBF) kernel SVM implementation[20]. This system could draw decision boundaries between data points from different classes and separate them with maximum margin[21]. The random forest fitted 100 decision trees with all extracted features. We used the bootstrap aggregation algorithm for creating multiple different models from a single training dataset.

**Statistical analysis**

We reported the descriptive summaries as mean ± standard deviation for continuous variables and as frequency (percentage and confidence interval) for categorical variables. We used accuracy, sensitivity, specificity to evaluate the performance of a predictive model. Receiver operating characteristic (ROC) curves, positive predictive value (PPV), and negative predictive value (NPV) were also obtained. All statistical tests used in this study were 2-sided, and a P value less than 0.05 was considered significant. The prediction of hemorrhage was presented by the ANN as a scaled value where outputs greater than 0.5 predicted in favor of hemorrhage and values less than 0.5 predicted the absence of hemorrhage.

## Results

A total of 378 patients were included in this study (figure 1). Table 1 lists the baseline characteristics between patients with hemorrhagic moyamoya disease and ischemic moyamoya disease. A total of 126 (33.3%) patients were hMMD and 252 (66.7%) were iMMD. We identified 17 of 126 hMMD patients (13.5%) combined with microaneurysms, whose proportion is significantly higher than that of the iMMD group (0.8%, P<0.001). Compared to iMMD, more PCoA or AChA dilation were detected in hMMD (87.3% vs. 83.3%, P=0.313). The majority of iMMD patients (139, 55.2%) had advanced Suzuki stage (stage >3) compared with hMMD patients (63, 50.0%, p = 0.343). We extracted 33 parameters, including 11 demographic and 22 radiographic features (Table 1, 2 and Supplementary Materials) as the input.

**ANN Model**

After randomization, 302 patients (80%) were enrolled in the training set, and 76 patients (20%) were enrolled in the test set. We set up a three-layer ANN architecture containing a single hidden layer with 100 hidden units (figure 2). Table 3 shows the performance of ANN model prediction on the test dataset after 1000 epochs model updates. The average accuracy for the prediction of hemorrhage was 75.7% (95% CI, 68.6%–82.8%) . The sensitivity and specificity were 67.3% (95% CI, 46.3% – 88.3%) and 79.8% (95% CI, 65.1% – 94.5% ), respectively. The PPV was 63.6% (95% CI, 50.5% – 76.7%), and the NPV was 83.4% (95% CI, 76.5% – 90.3%). The area under the curve (AUC) was 0.913 (figure 3).

**SVM and Random Forest Models**

Table 3 shows the prediction performance of all prediction models based on the test dataset. The accuracy of SVM and random forest were 69.2% (95% CI, 56.9% – 81.5%) and 70.0% (95% CI, 57.0% – 83.0%) , respectively. The sensitivity and specificity of SVM were 13.0% and 87.1%, respectively. For random forest, the sensitivity and specificity were 19.8% and 80.6%, respectively.

**Discussion**

In this study, we developed three machine learning models for hemorrhage risk prediction of MMD on the basis of initial presentation, medical history and radiographic features. To our knowledge, this is the first study that developed machine learning models to predict hemorrhage in moyamoya disease.

With rapid advances in computer science, AI is proving to be effective in most aspects of medicine, including diagnosis, planning and even treatment. Deep learning study in stroke medicine is rising in recent years. Pustina et al.[22] proposed an automated segmentation algorithm which could learn the relationship between existing manual segmentations and a single T1-weighted MRI. They developed an algorithm with a dataset of 60 left-hemispheric chronic stroke patients and tested with an independent dataset. And the CNN system achieved the satisfactory correlation between predicted lesion volume and manual lesion volume (r=0.957). Takahashi et al.[23] used a leave-one-case-out method to identify hyperdense middle cerebral artery (MCA) dot sign on CT. The system achieved a sensitivity of 97.5% for detection of the MCA dot sign at a false-positive rate of 0.5 per image. ANN is a data-driven neural network and an effective tool for modeling. If a sufficient amount of information is given, it is possible to obtain an output value with high predictive accuracy without having explicit knowledge about the internal subprocess[24,25]. Dumont et al.[26] designed an ANN to help predict symptomatic cerebral vasospasm in an adult population. Liu et al. developed a two-layer feed-forward ANN framework to predict anterior communicating artery rupture risk and get the overall prediction accuracy was 94.8 %[27].

Given the absence of randomized trials and guidelines, therapeutic decisions of moyamoya are largely based on clinicians' and surgeons' experience and center tools [1,28]. Surgical revascularization appears the most effective strategy to improve cerebral hemodynamics in MMD patients. Surgical techniques can be classified into three groups: direct, indirect, and combined revascularization procedures[29,30]. The direct bypass could improve cerebral hemodynamics immediately, although it could be challenging in pediatric patients. Indirect revascularization is technically straight-forward; however, the development of the collateral network and the increase in cerebral blood flow usually require 3–4 months[28]. A multicenter trial reported direct revascularization reduces the risk for recurrent hemorrhage in adult patients presenting with hemorrhage secondary to MMD[31].

The mechanism of hemorrhage among adult MMD patients is complex and predicting the probability of bleeding is still challenging. The risk factors for bleeding in MMD patients were studied previously, including demographic factors and radiographic factors. Previous studies have described that hMMD have a female dominance in sex distribution[2,32]. As for radiographic features, it was reported that 71.9% of hemorrhagic hemispheres from adult patients manifested AChA dilatation and branching, and dilatation and abnormal branching of the PCoA was relatively low in hemorrhagic hemispheres from adult patients (18.8%)[33]. Both of them were significantly higher than in the ischemic and asymptomatic hemispheres. Using dilatation and abnormal branching of the AChA and/or PCoA as predictors, Morioka et al. [33] obtained high specificity (86.4%) and sensitivity (84.4%) for hemorrhagic events in adult moyamoya patients. The results of the JAM trial provide strong support for the hypothesis that hemodynamic stress on the lenticulostriate collaterals is a frequent cause of hemorrhage in MMD[31]. Rupture of microaneurysms and abnormal vessels at the base of the skull were also considered as a possible cause. In this study, dilatation and abnormal branching of the anterior choroidal artery and/or posterior communicating artery could also be considered strong prognostic variables of hemorrhage[34].

These three machine learning models have reliable prediction accuracy. ANN achieved the highest accuracies among all compared classification algorithms in our study. The other two frameworks, SVM and random forest, had become popular machine learning algorithms owing to their less processing power requirement. However, it should be pointed out that SVM and random forest obtained poor performance in this study, considering lower sensitivities and PPVs suggesting more hemorrhage were not predicted. The less satisfied results of SVM and random forest might be attributed to relatively small sample size and imbalanced data.

The neural network systems require no previous knowledge of the data and could be optimized continuously as more input data were fed into the system.[35] The strength of the neural network is in its ability to learn from a large dataset and recognize patterns that can be used to predict future risks[36]. To our knowledge, no previous studies have compared the accuracy of ANN compared to regression analysis methods in the field of hemorrhage prediction in moyamoya disease. As our results have shown that the accuracy of a hidden layer ANN could be more reliable than the single predictor of PCoA and AChA enlargement in our cohort. The deep learning system could help physicians assess potential hemorrhagic risk in moyamoya disease, which may translate into useful clinical information in counseling for the necessity of revascularization.

**Limitation**

There are limitations to this study. The included data were from a single stroke center in China and the number of patients were small, thus, the algorithm may not perform as well for the generalized population with moyamoya disease. Validation of this algorithm on additional data sets is required.

**Conclusion**

In conclusion, machine learning systems like ANN could be used to predict the possibility of hemorrhage among adult MMD patients based on clinical information and radiographic features. These algorithms could assist physicians to discern possible hMMD patients in the future and put the interventional strategy into practice for better prognosis.

**Tables**

Table 1. The baseline characteristics between patients with hMMD and iMMD.

| Variables | Total | Hemorrhagic MMD (%) | Ischemic MMD (%) | P |
|---|---|---|---|---|
| Number of patients | 378 | 126 (33.3%) | 252 (66.7%) | |
| Age | 38.2 ± 9.39 | 38.8 ± 8.9 | 37.9 ± 9.6 | 0.374 |
| Females | 204 (54.0%) | 74 (58.7%) | 130 (51.6%) | 0.189 |
| Vascular risk factors | | | | |
|   Hypertension | 112 (29.6%) | 23 (18.3%) | 89 (35.3%) | 0.001 |
|   Diabetes | 26 (6.9%) | 3 (2.4%) | 23 (9.1%) | 0.015 |
|   Antiplatelet drugs | 17 (4.5%) | 3 (2.4%) | 14 (5.6%) | 0.16 |
|   Drinking History | 24 (6.3%) | 5 (4.0%) | 19 (7.5%) | 0.179 |
|   Smoking History | 30 (7.9%) | 9 (7.1%) | 21 (8.3) | 0.686 |
|   Hyperlipidemia | 11 (2.9%) | 1 (0.8%) | 10 (4.0%) | 0.16 |
|   Ischemic stroke history | 169 (44.7%) | 14 (11.1%) | 155 (61.5%) | <0.001 |
|   mRS on admission≥3 | 34 (9.0%) | 14 (11.1%) | 20 (7.9%) | 0.309 |

Abbreviations: mRS, modified rankin scale

Table 2. Key radiographic features of patients with hMMD and iMMD.

| Radiographic features | Total | Hemorrhagic MMD (%) | Ischemic MMD (%) | P |
|---|---|---|---|---|
| Unilateral moyamoya | 44 (11.6%) | 17 (13.5%) | 27 (10.7%) | 0.427 |
| With Aneurysm | 19 (5.0%) | 17 (13.5%) | 2 (0.8%) | <0.001 |
| Suzuki stage >3 | 202 (53.4%) | 63 (50.0%) | 139 (55.2%) | 0.343 |
| L-mma collaterals | 142 (37.6%) | 56 (44.4%) | 86 (34.1%) | 0.051 |
| R-mma collaterals | 144 (38.1%) | 51 (40.5%) | 93 (36.9%) | 0.5 |
| Acha/pcom dilation | 320 (84.7%) | 110 (87.3%) | 220 (83.3%) | 0.313 |

Abbreviations: acha/pcom dilation, dilation of either anterior choroidal artery and posterior communicating artery; L-mma collaterals, collateral arteries originated from left middle meningeal artery;L-mma collaterals, collateral arteries originated from left middle meningeal artery.

Table 3 Prediction accuracy of ANN, SVM and Random Forest.

|  | Accuracy (95% CI) | Sensitivity (95% CI) | Specificity (95% CI) | PPV (95% CI) | NPV (95% CI) |
|---|---|---|---|---|---|
| ANN | 75.7% (68.6%-82.8%) | 67.3% (46.3%-88.3%) | 79.8% (65.1%-94.5%) | 63.6% (50.5%-76.7%) | 83.4% (76.5%-90.3%) |
| SVM | 69.2% (56.9%-81.5%) | 13.0% (0%-33.4%) | 87.1% (69.0%-99.9%) | 31.6% (0%-67.8%) | 68.5% (64.4%-72.5%) |
| Random Forest | 70.0% (57.0%-83.0%) | 19.8% (0%-41.8%) | 80.6% (61.5%-99.7%) | 32.2% (5.6%-58.8%) | 68.5% (63.0%-74.0%) |

Abbreviations: ANN, artificial neural network; SVM, Support Vector Machine; PPV, positive predictive value; NPV, negative predictive value.

**Figures**

Figure 1: Flow diagram for the process of participant selection.

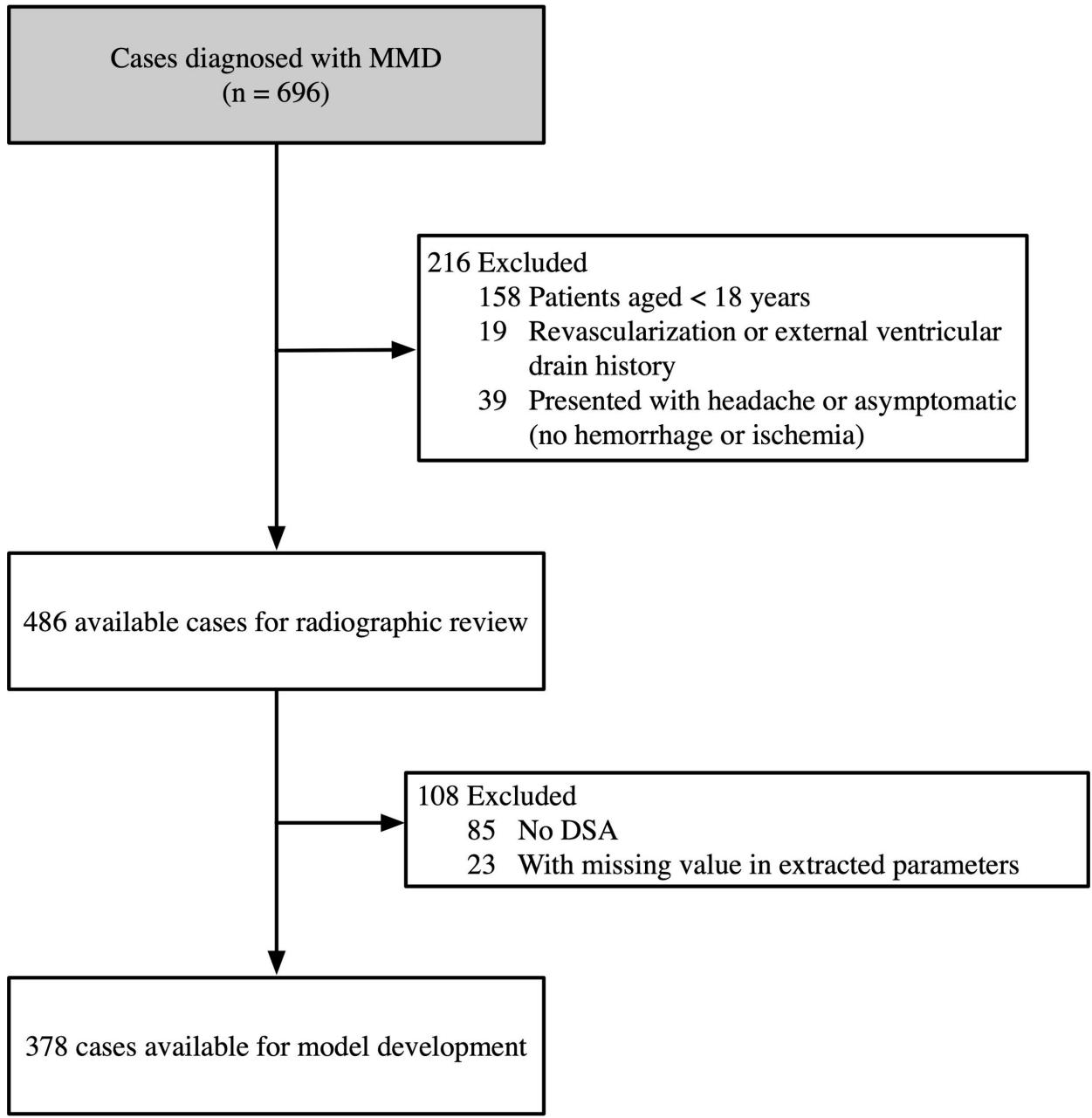

Figure 2: Plot of the ANN network graph.

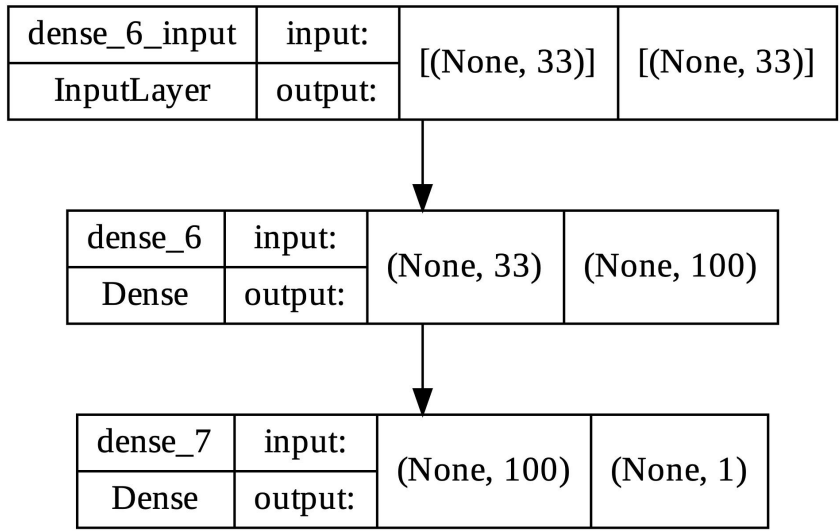

Figure 3: ROC curves of ANN on test datasets. The AUC is 0.913.

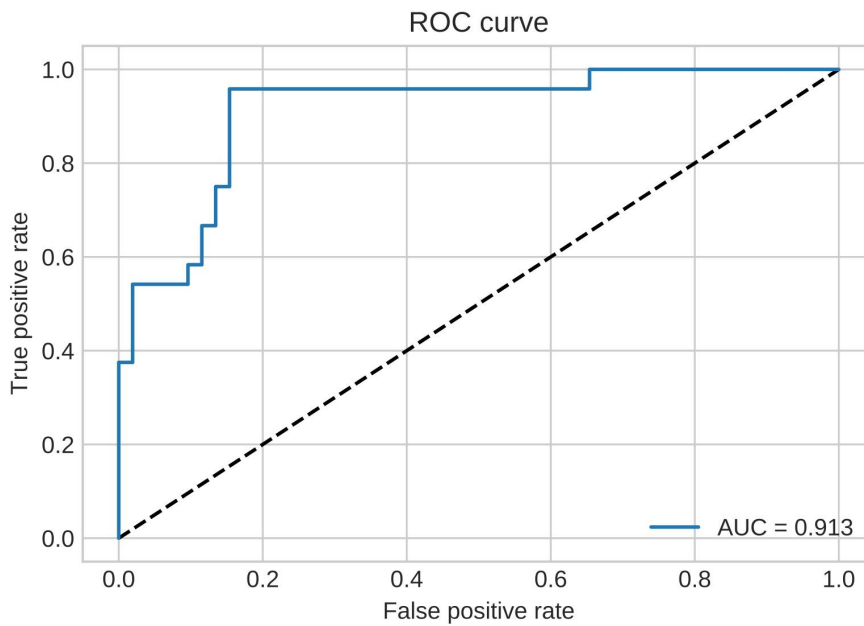

**Acknowledgments**
This study was funded by the program of the Beijing Municipal Commission of Education (KM201910025014, for Qian Zhang).

Deep Learning to Predict Hemorrhage in MMD                    15**Competing interests and Disclosures**
All authors declare no competing interests. All authors report no disclosures.